\documentclass[letterpaper]{article} 
\usepackage{aaai25}  
\usepackage{times}  
\usepackage{helvet}  
\usepackage{courier}  
\usepackage[hyphens]{url}  
\usepackage{graphicx} 
\usepackage{natbib}  
\usepackage{caption} 
\usepackage{algorithm}
\usepackage{algorithmic}
\usepackage{subfigure}
\usepackage{multirow}
\usepackage{graphicx}

\usepackage[utf8]{inputenc} 
\usepackage[T1]{fontenc}    
\usepackage{url}            
\usepackage{booktabs}       
\usepackage{amsfonts}       
\usepackage{nicefrac}       
\usepackage{microtype}      
\usepackage{xcolor}         
\usepackage{amsmath}
\usepackage{float}
\usepackage{makecell}
\usepackage{comment}
\usepackage{newfloat}
\usepackage{listings}
\DeclareCaptionStyle{ruled}{labelfont=normalfont,labelsep=colon,strut=off} 
\floatstyle{ruled}
\newfloat{listing}{tb}{lst}{}
\floatname{listing}{Listing}

\usepackage{colortbl}
\usepackage{threeparttable}

\newcommand{\boldres}[1]{{\textbf{{#1}}}}

\usepackage{amsmath,amsfonts,bm}

\def\eqref#1{equation~\ref{#1}}

\def\1{\bm{1}}

\DeclareMathAlphabet{\mathsfit}{\encodingdefault}{\sfdefault}{m}{sl}
\SetMathAlphabet{\mathsfit}{bold}{\encodingdefault}{\sfdefault}{bx}{n}

\title{Geolocation Representation from Large Language Models are Generic Enhancers for Spatio-Temporal Learning}
\author{
    Junlin He\equalcontrib ,
    Tong Nie\equalcontrib,
    Wei Ma\thanks{Corresponding author.}\\
}
\affiliations{
    The Hong Kong Polytechnic University \\

    Hong Kong SAR, China\\
    \{junlinspeed.he, tong.nie\}@connect.polyu.hk,
    wei.w.ma@polyu.edu.hk
}

\usepackage{bibentry}

\begin{document}

\maketitle

\begin{abstract}
In the geospatial domain, universal representation models are significantly less prevalent than their extensive use in natural language processing and computer vision. This discrepancy arises primarily from the high costs associated with the input of existing representation models, which often require street views and mobility data. 
To address this, we develop a novel, training-free method that leverages large language models (LLMs) and auxiliary map data from OpenStreetMap to derive geolocation representations (LLMGeovec). LLMGeovec can represent the geographic semantics of city, country, and global scales, which acts as a generic enhancer for spatio-temporal learning. Specifically, by direct feature concatenation, we introduce a simple yet effective paradigm for enhancing multiple spatio-temporal tasks including geographic prediction (GP), long-term time series forecasting (LTSF), and graph-based spatio-temporal forecasting (GSTF). 
LLMGeovec can seamlessly integrate into a wide spectrum of spatio-temporal learning models, providing immediate enhancements.
Experimental results demonstrate that LLMGeovec achieves global coverage and significantly boosts the performance of leading GP, LTSF, and GSTF models. Our codes are available at \url{https://github.com/Umaruchain/LLMGeovec}.


\end{abstract}

%

\begin{figure*}[!htbp]
  \centering
  \captionsetup{skip=1pt}
  \includegraphics[width=1\textwidth]{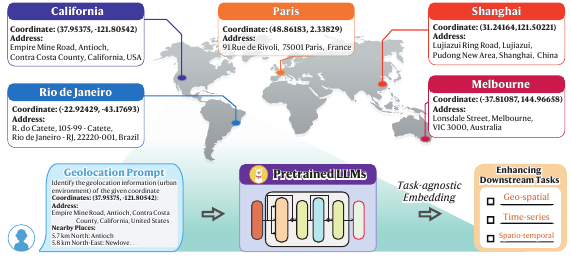}
  \caption{Our geolocation representation method consists of two phases: firstly, generating geolocation prompts for coordinates from map data, and then generating representations for text descriptions from pre-trained LLMs. It achieves global coverage and generates representations that can be used in various spatio-temporal tasks.}
  \label{fig:intro}
\end{figure*}

\section{Introduction}
Geolocation representation models encode geographical coordinates into latent embeddings with enriched geographic contextual information. 
Such embeddings ensure that similar representations reflect analogous sociodemographic attributes, activity patterns, and climatic characteristics across locations over the globe~\citep{wang2022self,jean2019tile2vec,lee2021predicting,wang2020urban2vec,zhang2021multi,zhang2023spatial,zhou2023hierarchical,kim2022effective}.

These geolocation representations are naturally suited to enhance spatio-temporal learning because they carry spatial contextual semantics. Previous research has focused only on how geolocation representations can be used for geographic prediction (GP). Specifically, GP tasks are trained on the attributes of some locations and used to predict the attributes of the remaining locations. 
These attributes include crime rate ~\cite{li2022spatial,kim2022effective}, poverty rate ~\cite{jean2016combining,chi2022microestimates,marty2024global}, public health  ~\citep{yeh2021sustainbench,nilsen2021review,draidi2022spatial,chang2022role,sheehan2019predicting}, and so on. However, these geolocation representations are not used for more complex tasks.

The reason why previous approaches do not further extend the complex applications is that they do not achieve global coverage and they have a heavy reliance on expensive input data such as street views, travel patterns, and traffic trajectories ~\citep{wang2020urban2vec,kim2022effective,lee2021predicting,zhang2023spatial}. Although studies have used free and globally available satellite imagery for geolocation representation, their effectiveness has been hampered by the low resolution and absence of important features such as activity patterns ~\citep{manvi2023geollm,robinson2017deep,head2017can,jean2019tile2vec,elmustafa2022understanding,xi2022beyond,logcan,sun2024ultra}.

Our objective is to develop a generic and effective geolocation representation method that utilizes only readily accessible global data to improve more spatio-temporal learning tasks: GP, long-term time series forecasting (LTSF), and graph-based spatio-temporal forecasting (GSTF). The latter two are typical spatio-temporal datasets that, given the values of many nodes at historical moments, predict the future values of those nodes. They differ in that the former tends to deal with correlations between nodes by channel-mixing strategies while the latter aggregates spatial connections between nodes by graph neural networks (GNNs).




Recent advancements have demonstrated the extensive spatio-temporal and human-related knowledge embedded within large language models (LLMs). Some studies have transformed GP tasks as text generation tasks in LLMs ~\citep{manvi2023geollm,manvi2024large}, and some have even found that LLMs learn linear representations of space and time across multiple scales (global, country, city) ~\citep{gurnee2023language}.
Inspired by these findings, we explore the potential of LLMs to generate effective geolocation representations.

In this paper, we introduce a novel, training-free method that uses LLM and OpenStreetMap auxiliary map data to derive geolocation representations (LLMGeovec). As illustrated in Fig. ~\ref{fig:intro}, our approach first extracts textual descriptions of the coordinates from OpenStreetMap, which provides a sufficient geographic context. LLMs process these descriptions, and the final hidden states of individual tokens are averaged to form the LLMGeovec embedding for each coordinate.

LLMGeovec is not only the first geolocation representation model to achieve global coverage using LLMs, it also presents a simple yet effective paradigm for enhancing spatio-temporal learning with LLMs. In GP, LLMGeovec can be used either as a standalone representation or concatenated with existing representations generated using street view and human mobility data. 
In LTSF, different temporal patterns of various nodes need to be handled, and the challenge is how to not only identify the unique characteristics of nodes but also model the correlations between them ~\citep{nie2024channel,nie2022time,zeng2022transformers}. LLMGeovec can be concatenated to the temporal features of individual nodes, allowing models to distinguish between different nodes while modeling their geographical connections. In terms of GSTF, current approaches focus on capturing spatial dependencies by using GNNs to aggregate node features with the guidance of an adjacency matrix~\citep{shao2022spatial,shao2022decoupled,shao2022pre,wu2019graph,chen2020multi,geng2019spatiotemporal,nie2023correlating}. Adding LLMGeovec as new node features before performing GNNs can provide more a priori knowledge about the spatial semantics of nodes.



Our experiments demonstrate that LLMGeovec is plug-and-play and improves the performance of various spatio-temporal learning. In GD, LLMGeovec alone achieves SOTA performance for tasks of all scales, and its performance is further enhanced when spliced with other features. 
In LTSF and GSTF, LLMGeovec enhances many of the latest temporal models and spatio-temporal graph neural networks (STGNNs). Notably, we find that LLMGeovec with a simple MLP can outperform many STGNNs, demonstrating that LLMGeovec already contains rich spatial correlations, and has the potential to replace heavy GNNs.

In summary, we present three major contributions:
\begin{itemize}
\item We propose LLMGeovec, a novel, training-free approach that leverages LLMs to generate semantically rich geolocation representations. By utilizing OpenStreetMap data, LLMGeovec functions as a universal and effective geolocation representation model.
\item LLMGeovec achieves comprehensive global geographic coverage and offers a simple yet effective paradigm for enhancing spatio-temporal learning using LLMs, resulting in direct performance improvements.
\item Extensive experimental analysis demonstrates that LLMGeovec achieves global coverage and significantly boosts the performance of leading GD, LTSF, and GSTF models.
\end{itemize}

\section{Related Work}
\subsection{Geolocation Representation Models}
Geolocation representation models encode spatial coordinates into latent embeddings enriched with contextual geographic information. These embeddings ensure that similar representations reflect analogous social attributes and climatic characteristics across diverse locations~\citep{wang2022self,jean2019tile2vec,lee2021predicting,wang2020urban2vec,zhang2021multi,zhang2023spatial,zhou2023hierarchical,kim2022effective}. 

Currently, there are three primary types of geolocation representation models: GNN-based models, image-based models, and natural language processing (NLP)-based models.
GNN-based models construct graphs from correlations between locations, such as geographic distance, points of interest (POI), and human mobility patterns. These models generate node representations through message passing on the constructed graphs~\citep{zhang2021multi,kim2022effective,zhang2023spatial,zhou2023hierarchical}. In contrast, Image-based models utilize street view or satellite imagery and employ contrastive learning to generate representations tied to specific coordinates~\citep{jean2019tile2vec,wang2020urban2vec,liu2023knowledge,li2022predicting,xi2022beyond,tinyvim}. Meanwhile, NLP-based models leverage textual representations to represent the textual descriptions associated with the corresponding locations.

However, GNN-based models are highly dependent on human mobility data, which restricts their applicability to urban environments where such records are available. This limitation also challenges the modeling of cities on a global scale. Image-based models face challenges as well; those relying solely on satellite imagery often lack critical human activity information ~\citep{xi2022beyond,manvi2023geollm}, while street view images can be costly and not universally accessible. NLP-based models show promise due to the abundance of geographically relevant textual data available online, which is typically free and globally accessible. However, existing NLP-based models, such as using Doc2vec to represent textual descriptions from Wikipedia, are inherently limited in their data source and model ability to fully capture the richness of geographic information ~\citep{sheehan2019predicting}.

To address these limitations, this paper introduces LLMGeovec, an NLP-based geolocation representation model that extracts extensive spatio-temporal and human-related knowledge compressed in LLMs to represent locations effectively.

\subsection{LLMs for GP}
Recent advancements in LLMs have seen their application in various GP tasks. Utilizing pre-trained LLMs, researchers have addressed various challenges such as forecasting dementia patterns over time series data, predicting urban functionalities, and estimating socio-climatic variables ~\citep{mai2023opportunities,manvi2024large,zhang2023geogpt}. Despite these applications, the efficacy of LLMs not specifically fine-tuned for geographic tasks remains suboptimal.
Significant efforts have been directed towards customizing LLMs for geospatial analytics. For example, some researchers have fine-tuned LLMs in geoscience text corpora to enhance their performance in geographic question answering, summarization, and text classification tasks ~\citep{deng2024k2}. Furthermore, more recent studies have constructed training sets derived from OpenStreetMap data and associated GP tasks, leading to improved performance by fine-tuning LLMs on training sets ~\citep{manvi2023geollm}. 

However, fine-tuning LLMs is resource-intensive, often requiring substantial computational and data resources ~\citep{hu2021lora,kaddour2023challenges}. 
Previous approaches mainly generate texts with LLMs to approximate GP, focusing more on relevance rather than precision ~\citep{lopez2023can}. In contrast, our proposed LLMGeovec framework leverages pre-trained LLMs for direct geolocation representation. This approach facilitates the use of geographic knowledge of LLMs within various prediction models. Our experimental results confirm the robustness and utility of LLMGeovec in practical GP.

\subsection{LLMs for LTSF and GSTF}

LTSF and GSTF involve the analysis of spatio-temporal data, which encapsulate both the temporal dynamics of individual nodes and the spatial dependencies among them. Recent advancements have explored the integration of LLMs to leverage their sequence modeling capabilities and the spatio-temporal knowledge they encode. This is typically achieved by tokenizing time series and graph data, fine-tuning LLMs on these tokens, and subsequently employing customized prompts to improve forecast accuracy \citep{jiang2024empowering, li2024urbangpt, zhou2023one, chang2024llm4ts, sun2023test, cao2023tempo, jin2023time}.

However, existing approaches predominantly utilize LLMs as direct predictors, necessitating substantial computational resources for fine-tuning and often falling short in embedding spatio-temporal knowledge into existing advanced forecasting models. Our work addresses this limitation by extracting geolocation representations from LLMs, enriching the LTSF and GSTF models with enhanced spatial correlation learning, leading to direct performance improvements.

\section{Preliminaries}
In this section, we introduce the definitions of geolocation representation learning, GP, LTSF and GSTF.

\textbf{Geolocation Representation Learning.}
Given a set of nodes \(P \in \mathbb{R}^{2 \times N}\), where \(N\) represents the number of nodes, and \( P_i = (P_{i}^{\text{lon}}\), \(P_{i}^{\text{lat}})\) denotes the longitude and latitude of the \(i\)-th node, the goal of geolocation representation learning is to construct an effective encoder \(f\) that transforms \(P\) into geographically informative representations \(Z = f(P) \in \mathbb{R}^{M \times N}\) with \(M\) denoting the dimension of the representation.

\textbf{GP.}
Given a set of nodes \(P \in \mathbb{R}^{2 \times N}\), each node is associated with geographic attributes such as climatic indicators (e.g., average annual temperature, humidity) and social indicators (e.g., regional average educational attainment, average annual income, poverty rate, crime rate). For a given set of geographic attributes \(A \in \mathbb{R}^{1 \times N}\), GP in the context of geolocation representation learning involves training a linear regressor with \(Z_{[:K]} = f(P_{[:K]})\) to fit \(A_{[:K]}\) using \(K\) training samples. The performance of the regressor on the test sets \(Z_{[K:]} = f(P_{[K:]})\) and \(A_{[K:]}\) is used to measure the quality of the encoder \(f\) and the representations $Z$.

\textbf{LTSF.}
We consider a multivariate time series (MTS) \(X \in \mathbb{R}^{H \times N}\), where \(N\) represents the number of nodes (variates) and \(H\) is the number of historical time slots. The objective is to predict future values \(Y \in \mathbb{R}^{F \times N}\), with \(F\) as the number of future time slots. Each value of node \(i\) in time slot \(t\) is denoted by \(X_t^i\), and their coordinates by \(P \in \mathbb{R}^{2 \times N}\).

\textbf{GSTF.}
Different from LTSF, GSTF constructs a weighted adjacency matrix \(A \in \mathbb{R}^{N \times N}\), where \(A_{ij} = 1/\text{dist}(P_i,P_j)\) and \(\text{dist}(P_i,P_j)\) represents the spatial distance between node $P_i$ and $P_j$ . A graph \(G\) is then formed based on \(A\). Unlike LTSF, GSTF leverages GNNs to aggregate the features of nodes \(X_t\) at $t$-th time slot, enhancing prediction by incorporating spatial relationships.

\section{LLMGeovec: A Generic Enhancer for Spatio-Temporal Learning}
As depicted in Fig. ~\ref{fig:intro}, the proposed LLMGeovec encapsulates two primary phases: prompt generation and text embedding via LLMs. Initially, we generate geographic descriptions based on specified coordinates with map data. These descriptions are then transformed into embeddings by LLMs. The obtained embeddings can be used for GP, LTSF, and GSTF.

\subsection{Prompt Generation}
Given a coordinate, we generate universal prompts, intentionally devoid of task-specific data, to enable the effective extraction of geographic knowledge of LLMs. As outlined in Fig. ~\ref{fig:intro}, the prompt structure incorporates:
\begin{itemize}
    \item \textbf{Instruction}: guides LLMs in identifying essential geographic information linked to specific coordinates.
    \item \textbf{Address}: uses reverse-geocoding to detail the hierarchy of location, from local neighborhoods to national identifiers.
    \item \textbf{Nearby Places}: enumerates the ten nearest points of interest within a 100-kilometer radius, including their names, distances, directions and bearings.
\end{itemize}
Data sources include OpenStreetMap \citep{neis2012analyzing}, with addresses derived through Nominatim's reverse geocoding \citep{serere2023enhanced} and nearby places via the Overpass API \citep{olbricht2011overpass}. This approach aligns with and extends previous studies \citep{manvi2023geollm,manvi2024large} by focusing on the extraction of generalized geographic information without specifying downstream tasks.

\subsection{Text Embedding Using LLMs}
With the geolocation prompts generated, we proceed to embed these textual descriptions using LLMs. Recent studies have explored enhancing text embeddings generated by LLMs, typically by modifying attention mechanisms or repeating prompts to circumvent the limitations of decoder-only models ~\citep{behnamghader2024llm2vec,muennighoff2022sgpt,ma2024fine,wang2023improving,springer2024repetition}. 
Our structured prompts, particularly with crucial geographic context presented at the end of prompts, allow LLMs to generate sufficiently high-quality geolocation representations without repetition of prompts or modification of models. To be specific, we use the average word embeddings from the last layer of a pre-trained LLM as the text representation, ensuring our LLMGeovec method remains adaptable to the latest LLMs without training. In addition, by avoiding fine-tuning, our method preserves the intrinsic geographic knowledge within LLMs ~\citep{zhai2023investigating,lin2023speciality}. 

\subsection{Incorporating LLMGeovec into GP}
Consistent with many previous studies, high-quality geolocation representations can be used for GP with the help of partial region labeling \citep{wang2022self,jean2019tile2vec,lee2021predicting,wang2020urban2vec,zhang2021multi,zhang2023spatial,zhou2023hierarchical,kim2022effective}. This is a direct application of LLMGeovec. Specifically, we divide the locations into a training set and a test set, and use linear regression in the training set to map geolocation representation to location attributes. This is followed by testing in the test set.
It is worth noting that since LLMGeovec achieves global coverage, it can be used in GPs of various scales (global, country, city) and can also be combined with other geolocation representations through feature concatenation.


\subsection{Incorporating LLMGeovec into LTSF}
\label{sec-MTS-with-llm}
In this section, we describe the integration of LLMGeovec with LTSF models. We start by outlining a general LTSF model, which typically consists of a token embedding layer $E$, an encoder $C$, and a predictor $D$ \citep{chen2023tsmixer,li2023revisiting,liu2023itransformer,yi2024frequency,zhang2023dfnet}. The embedding layer $E$ projects the the $t$-th historical record $X_t \in \mathbb{R}^{1 \times N}$ into hidden temporal embeddings $S_t = E(X_t) \in \mathbb{R}^{d_t \times N}$, where $d_t$ is the embedding dimension. Note that there can be a normalization operation in this embedder such as RevIN \cite{kim2021reversible} to address the nonstationarity of time series.
The encoder $C$ then models the node-to-node and slot-to-slot relationships across $H$ historical time slots, and the predictor $D$ generates predictions $\hat{Y} \in \mathbb{R}^{F \times N}$ for the future $F$ time slots. This process is formulated as follows:
\begin{equation}
    S = \{ S_0, \cdots, S_t, \cdots, S_H \}, \quad S_t = E(X_t),
\end{equation}
\begin{equation}
    \text{Loss}_{\text{LTSF}} = \min || D(C(S)) - Y ||_F^2,
\end{equation}
where the model parameters are updated automatically through gradient descent. In practice, the encoder $C$ can be instantiated by Transformer blocks, convolution, and MLPs, to model either channel dependencies or token correlations. Many state-of-the-art LTSF models follow this architectural template, such as TSMixer \cite{chen2023tsmixer}, RMLP \cite{li2023revisiting}, iTransformer \cite{liu2023itransformer}, and FreTS \cite{yi2024frequency}. For other Transformer-based models that employ token-wise embedding, such as models in \cite{wu2021autoformer,zhou2021informer,zhang2024skip}, we can adapt them with a simple inverting strategy \cite{liu2023itransformer}.

For an LTSF task, we collect the latitude and longitude of each node that generates the time series to construct the node set $P$. We then select an LLM (e.g., LLaMa3) and use our proposed LLMGeovec to generate the geolocation representation $Z \in \mathbb{R}^{M \times N}$. A two-layer MLP acts as an adapter for LLMGeovec, projecting $Z$ into a low-dimensional space $Z' = \text{Adapter}(Z) \in \mathbb{R}^{d_{s} \times N}$ to align with the LTSF task. This process is described by the following equations:
\begin{equation}
    Z' =  \text{Adapter}(Z),
\end{equation}
\begin{equation}
    S' = \{ S_0', \cdots, S_t', \cdots, S_H' \}, \quad S_t' = \text{Concat}(E(X_t), Z'),
\end{equation}
\begin{equation}
    \text{Loss}_{\text{LTSF}}' = \min || D(C(S')) - Y ||_F^2,
\end{equation}
where $S_i' \in \mathbb{R}^{(d_t + d_s) \times N}$ is formed by concatenating the series embeddings $E(X_i)$ with the geolocation representations $Z'$ along the feature dimension. The parameters of both the adapter and the original components are updated automatically via gradient descent.


\subsection{Incorporating LLMGeovec into GSTF}
\label{sec-GNN-with-llm}
Previous spatio-temporal prediction models often employ GNNs to capture spatial relationships between nodes, aggregating them into node features, which are then input into the temporal modeling component sequentially or alternatively \citep{shao2022spatial,shao2022decoupled,shao2022pre,wu2019graph,tang2022activity,tang2022domain,zhang2025thatsn}. We refer to the spatio-temporal model as $\text{STGNN}$. Given the graph $G$, the structural template is framed as follows:
\begin{equation}
\begin{aligned}
    &S = \{ S_0, \cdots, S_t, \cdots, S_H \}, \quad S_t = E(X_t),  \\
    &\hat{S} = \text{STGNN}(S, G), \\
    & \text{Loss}_{\text{GSTF}} = \min || D(\hat{S}) - Y ||_F^2. \\
\end{aligned}
\end{equation}
where the embedder $E$ projects the node signal to hidden states, and all node states are collected into the $\text{STGNN}$ processor to generate the graph representation $\hat{S}$. 


On top of them, we first concatenate LLMGeovec into node features (e.g., temporal readings), which are then processed by $\text{STGNN}$. This process is described by:
\begin{equation}
    Z' =  \text{Adapter}(Z),
\end{equation}
\begin{equation}
    S'  = \{ S_0', \cdots, S_t', \cdots, S_H' \}, \quad S_t' = \text{Concat}(E(X_t), Z'),
\end{equation}
\begin{equation}\label{eq:stgnn}
    \hat{S}' = \text{STGNN}(S', G), 
\end{equation}
\begin{equation}
    \text{Loss}_{\text{GSTF}}' = \min || D(\hat{S}') - Y ||_F^2,
\end{equation}
where the parameters of both the adapter, the STGNN and $D$ are updated automatically via gradient descent.

Note that this scheme is applicable for STGNNs with different types of processing methods. Specifically, different models have different instantiations of Eq. \ref{eq:stgnn}, e.g., spatio-temporal message passing mechanisms. Both the mainstream time-then-space and time-and-space STGNN family discussed in \cite{cini2023taming} can be seamlessly adopted simply by concatenating LLMGeovec into the input of each node.

\begin{table}[!htbp]
\caption{A multi-scale and multi-topic GP benchmark}
\label{Table: benchmark}
\centering
    \renewcommand{\arraystretch}{1.1} 
    \setlength{\tabcolsep}{1pt}
    \resizebox{\linewidth}{!}{
    \begin{tabular}{l|ccc|c}
    \toprule
    \textbf{Tasks} & \textbf{Source} & \textbf{Scale} & \textbf{Attribute} & \makecell[c]{\textbf{Training/} \\ \textbf{Testing}} \\
    \midrule
    Annual Air Temperature                          & Chelsa   & Global  & Climate   & 80k/20k                  \\
    Annual Precipitation                            & Chelsa   & Global  & Climate   & 80k/20k                  \\
    Monthly Climate Moisture                        & Chelsa   & Global  & Climate   & 40k/20k                  \\
    Population Density                              & WorldPop & Global  & Society   & 80k/20k                  \\
    Nighttime Light Intensity                       & EOG      & Global  & Society   & 80k/20k                  \\
    Human Modification Terrestrial                  & SEDAC    & Global  & Society   & 80k/20k                  \\
    Global Gridded Relative Deprivation             & SEDAC    & Global  & Society   & 80k/20k                  \\
    Ratio of Built-up Area to Non-built Up Area & SEDAC    & Global  & Society   & 80k/20k                  \\
    Child Dependency Ratio                          & SEDAC    & Global  & Society   & 80k/20k                  \\
    Subnational Human Development                   & SEDAC    & Global  & Society   & 80k/20k                  \\
    Infant Mortality Rates                          & SEDAC    & Global  & Society   & 80k/20k                  \\
    Asset Index                                     & DHS      & Global  & Society   & 20k/5k                   \\
    Sanitation Index                                & DHS      & Global  & Society   & 20k/5k                   \\
    Women BMI                                       & DHS      & Global  & Society   & 40k/10k                  \\
    \midrule
    Poverty Rate                                    & DHS      & Country & Society   & 5k/1k                    \\
    Population Density                              & FaceBook & Country & Society   & 5k/1k                    \\
    Women BMI                                       & DHS      & Country & Society   & 5k/1k                    \\
    \midrule
    Population Density                              & NYC      & City    & Society   & 1k/424                   \\
    Education Level                                 & NYC      & City    & Society   & 1k/424                   \\
    Income Level                                    & NYC      & City    & Society   & 1k/424                \\
    Crime Rate                                      & NYC      & City    & Society   & 1k/424                \\
    \bottomrule
    \end{tabular}
    }
\end{table}

\begin{table}[!htbp]
\caption{Baseline methods of GP}
\label{Table: baseline of GP}
\centering  
\resizebox{\linewidth}{!}{
\begin{tabular}{>{\raggedright\arraybackslash}p{5cm}|c|c}
\toprule 
\textbf{Methods} & \textbf{Scales} & \textbf{Data Resources} \\ 
\midrule 
Node2vec, GCN, GAT & City & Road Graph \\
\addlinespace  
ZE-Mob, MGFN, MV-PN & City & Road Graph, Mobility \\
HDGE, HUGAT, MVURE, HKGL & City & Check-in, PoI    \\
\addlinespace
Image Supervised Learning & Country & Street View \\
Object Counts & Country & Street View \\
Mapillarygcn & Country & Street View \\
\addlinespace
Bert-whitening & All & Map \\
GTE-large & All & Map \\
GTE-qwen2 7B & All & Map \\
LLMGeovec & All & Map \\
\bottomrule 
\end{tabular}
}
\end{table}

\begin{table*}[!htbp]
  \caption{Results of the LTSF benchmarks. We report the average forecast error of different models under different prediction lengths. Full results are presented in Appendix. Part of the baseline results are from \citep{nie2024channel}. IMP shows the average percentage of MSE improvement of LLMGeovec.}
  \label{tab:main}
  \vskip -0.0in
  \vspace{0pt}
  \renewcommand{\arraystretch}{1} 
  \centering
  \resizebox{1\textwidth}{!}{
  \begin{threeparttable}
  \begin{small}
  \renewcommand{\multirowsetup}{\centering}
  \setlength{\tabcolsep}{5pt}
  \begin{tabular}{cc|cc>{\columncolor{gray!20}}c>{\columncolor{gray!20}}c|cc>{\columncolor{gray!20}}c>{\columncolor{gray!20}}c|cc>{\columncolor{gray!20}}c>{\columncolor{gray!20}}c|cc>{\columncolor{gray!20}}c>{\columncolor{gray!20}}c|c}
    \toprule
    \multicolumn{2}{c}{\multirow{2}{*}{Models}}  &
    \multicolumn{2}{|c}{\rotatebox{0}{{iTransformer}}} &
    \multicolumn{2}{c|}{\rotatebox{0}{{w/ LLMGeovec}}} &
    \multicolumn{2}{c}{\rotatebox{0}{{TSMixer}}} &
    \multicolumn{2}{c|}{\rotatebox{0}{{w/ LLMGeovec}}} &
    \multicolumn{2}{c}{\rotatebox{0}{{RMLP}}}&
    \multicolumn{2}{c}{\rotatebox{0}{{w/ LLMGeovec}}} &
    \multicolumn{2}{|c}{\rotatebox{0}{{Informer}}} &
    \multicolumn{2}{c|}{\rotatebox{0}{{w/ LLMGeovec}}} &
    \multicolumn{1}{c}{\rotatebox{0}{IMP}}\\
    \cmidrule(lr){3-19} 
    \multicolumn{2}{c}{Metric}   & \multicolumn{1}{|c}{{MSE}} & \multicolumn{1}{c}{MAE}  & \multicolumn{1}{c}{MSE} & \multicolumn{1}{c|}{MAE}  & \multicolumn{1}{c}{MSE} & \multicolumn{1}{c}{MAE}  & \multicolumn{1}{c}{MSE} & \multicolumn{1}{c|}{MAE}  & \multicolumn{1}{c}{MSE} & \multicolumn{1}{c}{MAE}  & \multicolumn{1}{c}{MSE} & \multicolumn{1}{c|}{MAE} & \multicolumn{1}{c}{MSE} & \multicolumn{1}{c}{MAE} & \multicolumn{1}{c}{MSE} & \multicolumn{1}{c|}{MAE}  & $\%$\\
    \toprule
    {{{Global Wind}}} 
    &   &{4.582}  &{1.51}  &{\underline{\boldres{3.979}}}  &{\underline{\boldres{1.380}}}  &{4.261}  &{1.424}  &{\boldres{4.132}}  &{\boldres{1.407}}  &{4.905}  &{1.498}  &{\boldres{4.180}}  &{\boldres{1.414}}  &{4.905}  &{1.576}  &{\boldres{4.844}}  &{\boldres{1.566}}  &{13.30\%}   
    \\
    \midrule
    {{{Global Temp}}} 
    &   &{13.079}  &{2.653}  &{\boldres{11.945}}  &{\boldres{2.601}}  &{12.035}  &{2.480}  &{\underline{\boldres{11.441}}}  &{\underline{\boldres{2.398}}}  &{13.447}  &{2.558}  &{\boldres{12.525}}  &{\boldres{2.480}}  &{\boldres{18.370}}  &{\boldres{3.209}}  &{18.639}  &{3.234}  &{5.19\%}   
    \\
    \midrule
    {{{Solar Energy}}} 
    &   &{0.233}  &{0.262}  &{\underline{\boldres{0.206}}}  &{\underline{\boldres{0.265}}}  &{0.255}  &{0.294}  &{\boldres{0.219}}  &{\boldres{0.289}}  &{0.261}  &{0.313}  &{\boldres{0.235}}  &{\boldres{0.286}}  &{0.264}  &{\boldres{0.308}}  &{\boldres{0.263}}  &{0.313}  &{11.59\%}
    \\
    \midrule
    \multirow{1}{*}{{{Demand-SH}}} 
    &  & {0.331}  &{0.298}  &{\boldres{0.322}}  &{\boldres{0.297}}  &{0.355}  &{0.332}  &{\boldres{0.336}}  &{\boldres{0.305}}  &{0.345}  &{0.326}  &{\underline{\boldres{0.318}}}  &{\underline{\boldres{0.286}}}  &{0.896}  &{\boldres{0.618}}  &{\boldres{0.779}}  &{0.666}  &{2.47\%}  
    \\
    \midrule
    \multirow{1}{*}{{{Air Quality}}} 
    &   &{1.922}  &{0.631}  &{\boldres{1.856}}  &{\boldres{0.619}}  &{2.068}  &{0.665}  &{\boldres{1.989}}  &{\boldres{0.650}}  &{1.857}  &{0.627}  &{\underline{\boldres{1.820}}}  &{\underline{\boldres{0.613}}}  &{3.584}  &{0.864}  &{\boldres{2.858}}  &{\boldres{0.771}}  &{3.46\%} 
    \\
    \midrule
    \multirow{1}{*}{{{Traffic-SD}}} 
    &   &{0.136}  &{0.225}  &{\boldres{0.106}}  &{\boldres{0.201}}  &{0.116}  &{0.212}  &{\underline{\boldres{0.105}}}  &{\underline{\boldres{0.197}}}  &{0.205}  &{0.296}  &{\boldres{0.168}}  &{\boldres{0.264}}  &{0.199}  &{0.298}  &{\boldres{0.152}}  &{\boldres{0.254}}  &{22.01\%}   
    \\
    \bottomrule
  \end{tabular}
    \end{small}
  \end{threeparttable}
}
\end{table*}

\section{Numeric Experiments}
We study the effectiveness of LLMGeovec through extensive experiments. 
We first demonstrate that in geolocation representation models, LLMGeovec performs SoTA in GP tasks at all three scales: city, national, and global, and even outperforms end-to-end supervised training models. 
We examine two LLMs, LLaMa3 8B and Mistral 8x7B, both of which are able to produce high-quality geolocation representations with LLMGeovec.
Moreover, LLMGeovec is seamlessly embedded into various LTSF and GSTF models and directly improves the model performance under various tasks. Notably, in the GSTF tasks, the use of a simple MLP and LLMGeovec outperforms many GNN-based approaches, and LLMGeovec shows great potential as an alternative to time-consuming GNNs.
For a detailed description of the models and the datasets (GP, LTSF, GSTF), please refer to Appendix. 

\subsection{LLMGeovec for GP}
To comprehensively validate the quality of LLMGeovec and its effectiveness in GP tasks, we constructed a multi-scale, multi-topic benchmark encompassing a range of scenarios from city-level poverty rates to global population density. Unlike many existing powerful baselines, our approach can generate high-quality geolocation representations for any location without expensive data or extensive training.

\textbf{A Multi-scale and Multi-topic GP Benchmark.} As illustrated in Table~\ref{Table: benchmark}, at the global scale, we collect 14 GP tasks. These include three climate indicators such as Annual Air Temperature and 11 social indicators like Population Density and Human Modification (detailed descriptions please refer to Appendix). 
We use 100,000 locations with global coverage, generated by ~\citet{manvi2024large} (Africa: 19,855; Asia: 55,893; Europe: 6,825; North America: 8,440; South America: 5,189; Oceania: 2,049). 
In line with ~\citet{manvi2023geollm, manvi2024large}, each GP task is associated with a corresponding GeoTIFF file. 
For each coordinate, the average value of 12 pixels surrounding the coordinate is taken as the value of the coordinate. 
Following the protocol of ~\citet{kim2022effective}, we perform five cross-validation using ridge linear regression implemented in Sklearn ~\citep{feurer2020auto, mcdonald2009ridge}, and the average of mean absolute error (MAE), root mean square error (RMSE) and $R^2$ are reported. On the country and city scales, we use existing benchmarks, including social indicators in India  ~\citep{lee2021predicting} and NYC  ~\citep{zhou2023hierarchical}. We also employ ridge linear regression in Sklearn and report MAE, RMSE, and $R^2$ on test sets.

\textbf{Baselines.}
We compare the LLMGeovec generated by LLaMa3 8B ~\citep{touvron2023llama} and Mistral 8x7B ~\citep{jiang2023Mistral}. As shown in Tab. ~\ref{Table: baseline of GP}, we also compare the text embedding generated by Bert-whitening, GTE-large, and GTE-qwen2 7B ~\citep{su2021whitening,li2023towards}.  In the city and country scales, we additionally compare Image-based and GNN-based geolocation representation models.

\begin{table}[htbp]
\caption{Performance of GP (global)}
\label{Table: performance of GP (global)}
\centering
\begin{small}
    \renewcommand{\arraystretch}{1.2} 
    \setlength{\tabcolsep}{4pt}
    \resizebox{0.48\textwidth}{!}{
    \begin{tabular}{l|ccc|ccc|ccc|ccc|ccc}
    \toprule
    \multirow{2}{*}{\textbf{Tasks}} & \multicolumn{3}{c|}{LLMGeovec (LLaMa 3 8B)} & \multicolumn{3}{c|}{LLMGeovec (Mistral 8 x 7B)} & \multicolumn{3}{c|}{Bert-whitening (Bert base)} & \multicolumn{3}{c|}{GTE-large} & \multicolumn{3}{c}{GTE-qwen2 7B} \\
    \cmidrule{2-16} 
    & MAE & RMSE & R$^2$ & MAE & RMSE & R$^2$ & MAE & RMSE & R$^2$ & MAE & RMSE & R$^2$ & MAE & RMSE & R$^2$ \\
    \midrule
    Annual Air Temperature & \textbf{9.90} & \textbf{14.32} & \textbf{0.95} & 11.05 & 16.03 & 0.94 & 24.03 & 32.73 & 0.76 & 22.23 & 30.15 & 0.80 & 14.05 & 19.99 & 0.91 \\
    Annual Precipitation & \textbf{2016.60} & \textbf{3021.76} & \textbf{0.86} & 2176.82 & 3245.12 & 0.83 & 3717.73 & 5293.82 & 0.56 & 3519.57 & 5012.55 & 0.61 & 2604.86 & 3931.23 & 0.76 \\
    Monthly Climate Moisture & \textbf{1302.14} & \textbf{2021.21} & \textbf{0.55} & 1345.82 & 2097.32 & 0.52 & 1644.56 & 2715.44 & 0.19 & 1619.65 & 2610.16 & 0.25 & 1394.10 & 2288.89 & 0.43 \\
    Population Density & \textbf{695.10} & \textbf{1020.11} & \textbf{0.85} & 759.81 & 1115.38 & 0.82 & 1342.98 & 2185.67 & 0.30 & 1266.82 & 1986.15 & 0.42 & 896.52 & 1417.02 & 0.70 \\
    Nighttime Light Intensity & \textbf{3.55} & \textbf{4.58} & \textbf{0.97} & 3.79 & 4.89 & 0.96 & 8.55 & 11.37 & 0.81 & 7.63 & 9.99 & 0.85 & 4.77 & 6.15 & 0.94 \\
    Human Modification Terrestrial & \textbf{0.07} & \textbf{0.09} & \textbf{0.78} & 0.07 & 0.09 & 0.75 & 0.12 & 0.15 & 0.39 & 0.11 & 0.14 & 0.47 & 0.08 & 0.11 & 0.68 \\
    Global Gridded Relative Deprivation & \textbf{6.56} & \textbf{8.98} & \textbf{0.85} & 6.70 & 9.15 & 0.84 & 10.43 & 13.60 & 0.65 & 9.83 & 12.95 & 0.68 & 8.13 & 10.86 & 0.78 \\
    Ratio of Built-up Area to Non-built Up Area & \textbf{8.44} & \textbf{11.07} & \textbf{0.78} & 8.72 & 11.41 & 0.77 & 13.04 & 16.40 & 0.52 & 12.51 & 15.81 & 0.56 & 10.48 & 13.41 & 0.68 \\
    Child Dependency Ratio & \textbf{5.84} & \textbf{8.29} & \textbf{0.86} & 5.88 & 8.34 & 0.88 & 9.50 & 13.12 & 0.64 & 9.11 & 12.47 & 0.68 & 7.17 & 10.09 & 0.79 \\
    Subnational Human Development & \textbf{5.79} & \textbf{8.20} & \textbf{0.89} & 5.82 & 8.22 & 0.89 & 9.81 & 13.30 & 0.70 & 9.15 & 12.36 & 0.75 & 7.10 & 9.95 & 0.83 \\
    Infant Mortality Rates & \textbf{3.98} & \textbf{6.06} & \textbf{0.93} & 4.02 & 6.14 & 0.93 & 7.42 & 10.76 & 0.77 & 7.24 & 10.20 & 0.80 & 4.97 & 7.50 & 0.89 \\
    Asset Index & \textbf{0.02} & \textbf{0.03} & \textbf{0.93} & 0.02 & 0.03 & 0.92 & 0.06 & 0.08 & 0.53 & 0.05 & 0.07 & 0.62 & 0.04 & 0.06 & 0.78 \\
    Sanitation Index & \textbf{0.09} & \textbf{0.12} & \textbf{0.95} & 0.10 & 0.13 & 0.93 & 0.23 & 0.30 & 0.67 & 0.20 & 0.26 & 0.75 & 0.15 & 0.20 & 0.85 \\
    Women BMI & \textbf{0.76} & \textbf{1.01} & \textbf{0.95} & 0.83 & 1.12 & 0.94 & 1.82 & 2.38 & 0.77 & 1.55 & 2.04 & 0.83 & 1.16 & 1.57 & 0.90 \\
    \bottomrule
    \end{tabular}
    }
\end{small}
\end{table}

\begin{table}[htbp]
\caption{Performance of GP (country)}
\label{Table: performance of GP (country)}
\centering
\begin{small}
    \renewcommand{\arraystretch}{1.2} 
    \setlength{\tabcolsep}{10pt}
    \resizebox{1\columnwidth}{!}{
    \begin{tabular}{l|c|c|c}
    \toprule
    \multirow{2}{*}{Methods} & \multicolumn{3}{c}{$R^2$} \\
    \cmidrule{2-4}
                             & Poverty Rate & Population Density & Women BMI \\
    \midrule
    Image Supervised Learning & 0.51 & 0.85 & 0.52 \\
    Object Counts             & 0.52 & 0.81 & 0.53 \\
    Mapillarygcn              & 0.53 & 0.89 & 0.56 \\ \hline
    Bert-whitening (Bert base)& 0.51 & 0.77 & 0.45 \\ 
    LLMGeovec (Mistral 8 x 7B)& \textbf{0.66} & \textbf{0.96} & \textbf{0.65} \\
    LLMGeovec (LLaMa 3 8B)    & \textbf{0.66} & \textbf{0.96} & \textbf{0.65} \\
    \bottomrule
    \end{tabular}
    }
\end{small}

\end{table}

\textbf{Performances of GP.} As illustrated in Table ~\ref{Table: performance of GP (global)} ~\ref{Table: performance of GP (country)}, and  ~\ref{Table: performance of GP (city)}, the LLMGeovec family significantly outperforms Bert-whitening in generating geolocation representations from the same textual descriptions. This highlights the superiority of LLMs in leveraging a vast Internet corpus for pre-training. Additionally, the results show that LLaMa3 8B generally performs better than Mistral 8x7B. We attribute this to the multilingual corpus used in the pre-training of LLaMa3 8B, which enhances its understanding of geographic knowledge across different regions. At a global scale, LLMGeovec (LLaMa3 8B) demonstrates robust performance, with $R^2$ values exceeding 0.75 for all tasks except for Monthly Climate Moisture, which has an R2 of 0.55. Most tasks achieve $R^2$ values above 0.90. At the national scale, LLMGeovec (LLaMa3 8B) improves the $R^2$ scores across various tasks by 0.07 to 0.10, compared to the SOTA model (MapillaryGCN), which employs end-to-end supervised training using Street View data and GNNs. At the city scale, LLMGeovec outperforms or is comparable to baselines that utilize extensive human activity data and sophisticated graph learning techniques for all tasks except Crime Rate. Notably, concatenating the geolocation representations generated by LLMGeovec with those from existing methods (e.g., HKGL) substantially boosts the performance of downstream tasks.

\begin{table}[htbp]
\caption{Performance of GP (city)}
\label{Table: performance of GP (city)}
\centering
\begin{small}
\renewcommand{\arraystretch}{1.2} 
\setlength{\tabcolsep}{4pt}
\resizebox{0.49\textwidth}{!}{
\begin{tabular}{@{}lcccccccccccc@{}}
\toprule

\multirow{2}{*}{Methods} & \multicolumn{3}{c}{Poverty Rate} & \multicolumn{3}{c}{Education Level} & \multicolumn{3}{c}{Income Level} & \multicolumn{3}{c}{Crime Rate} \\
\cmidrule(lr){2-4} \cmidrule(lr){5-7} \cmidrule(lr){8-10} \cmidrule(lr){11-13}
& MAE & RMSE & R$^2$ & MAE & RMSE & R$^2$ & MAE & RMSE & R$^2$ & MAE & RMSE & R$^2$ \\
\midrule
Node2vec & 0.45 & 0.66 & 0.078 & 0.09 & 0.12 & 0.68 & 0.23 & 0.30 & 0.51 & 0.50 & 0.64 & 0.43 \\
GCN & 0.45 & 0.66 & 0.07 & 0.09 & 0.12 & 0.68 & 0.21 & 0.28 & 0.56 & 0.44 & 0.58 & 0.53 \\
GAT & 0.44 & 0.64 & 0.12 & 0.09 & 0.12 & 0.66 & 0.23 & 0.29 & 0.52 & 0.48 & 0.61 & 0.47 \\
ZE-Mob & 0.47 & 0.68 & 0.018 & 0.13 & 0.16 & 0.42 & 0.29 & 0.37 & 0.24 & 0.60 & 0.75 & 0.20 \\
MGFN & 0.45 & 0.66 & 0.07 & 0.11 & 0.14 & 0.56 & 0.25 & 0.33 & 0.41 & 0.50 & 0.64 & 0.42 \\
MV-PN & 0.46 & 0.65 & 0.11 & 0.14 & 0.18 & 0.26 & 0.32 & 0.40 & 0.12 & 0.63 & 0.77 & 0.16 \\
HDGE & 0.45 & 0.66 & 0.07 & 0.10 & 0.14 & 0.58 & 0.24 & 0.32 & 0.45 & 0.53 & 0.67 & 0.37 \\
HUGAT & 0.47 & 0.67 & 0.04 & 0.13 & 0.16 & 0.38 & 0.30 & 0.38 & 0.22 & 0.53 & 0.67 & 0.37 \\
MVURE & 0.45 & 0.65 & 0.12 & 0.10 & 0.12 & 0.66 & 0.23 & 0.29 & 0.52 & 0.47 & 0.61 & 0.48 \\
HKGL & 0.42 & 0.62 & 0.20 & 0.08 & 0.11 & 0.73 & 0.20 & 0.27 & 0.60 & 0.40 & 0.51 & 0.64 \\ \hline
Bert-whitening (Bert base) & 0.47 & 0.68 & 0.02 & 0.11 & 0.15 & 0.48 & 0.27 & 0.35 & 0.33 & 0.59 & 0.75 & 0.20 \\ 
LLMGeovec (LLaMa 3 8B) & 0.43 & 0.64 & 0.13 & 0.08 & 0.11 & 0.74 & 0.20 & 0.26 & 0.62 & 0.51 & 0.66 & 0.39 \\
\textbf{HKGL w/ LLMGeovec (LLaMa 3 8B)} & \textbf{0.39} & \textbf{0.58} & \textbf{0.28} & \textbf{0.08} & \textbf{0.10} & \textbf{0.76} & \textbf{0.19} & \textbf{0.25} & \textbf{0.652} & \textbf{0.39} & \textbf{0.50} & \textbf{0.64} \\
\bottomrule
\end{tabular}
}
\end{small}
\end{table}

\textbf{Performance of Prompt Variants.}
To examine the effect of the various parts of the prompts on the quality of LLMGeovec, we try several variants of the prompts and test them on GP tasks in NYC.
As shown in Tab. ~\ref{Table: performance of Various Prompts}, Address, which is the latitude and longitude counterpart of the location hierarchy, from local neighborhoods to national identifiers, is the most important part of motivating geographic knowledge in LLM. As for the K nearest places, we discover that when K is too small, it can not provide effective geographic information; when K is too large, the neighboring nodes will be too convergent, which leads to poor results in downstream tasks.

\begin{table}[htbp]
\caption{Performance of Prompt Variants}
\label{Table: performance of Various Prompts}
\centering
\begin{small}
\renewcommand{\arraystretch}{1.2} 
\setlength{\tabcolsep}{4pt}
\resizebox{0.49\textwidth}{!}{
\begin{tabular}{@{}lcccccccccccc@{}}
\toprule
\multirow{2}{*}{Prompt} & \multicolumn{3}{c}{Poverty Rate} & \multicolumn{3}{c}{Education Level} & \multicolumn{3}{c}{Income Level} & \multicolumn{3}{c}{Crime Rate} \\
\cmidrule(lr){2-4} \cmidrule(lr){5-7} \cmidrule(lr){8-10} \cmidrule(lr){11-13}
& MAE & RMSE & R$^2$ & MAE & RMSE & R$^2$ & MAE & RMSE & R$^2$ & MAE & RMSE & R$^2$ \\
\midrule
Instruction + Address + Top 10 NearbyPlaces & 0.13 & 0.74 & 0.62 & 0.39 & 0.17 & 0.75 & 0.61 & 0.36 & 0.12 & 0.65 & 0.53 & 0.34 \\
Instruction + Address + Top 5 NearbyPlaces  & 0.17 & 0.75 & 0.61 & 0.36 & 0.12 & 0.65 & 0.53 & 0.34 & 0.06 & 0.61 & 0.48 & 0.32 \\
Instruction + Address + Top 1 NearbyPlaces  & 0.12 & 0.65 & 0.53 & 0.34 & 0.06 & 0.61 & 0.48 & 0.32 & -0.01 & 0.42 & 0.29 & 0.09 \\
Instruction + Address & 0.06 & 0.61 & 0.48 & 0.32 & -0.01 & 0.42 & 0.29 & 0.09 & 0.13 & 0.74 & 0.62 & 0.39 \\
Instruction & -0.01 & 0.42 & 0.29 & 0.09 & 0.13 & 0.74 & 0.62 & 0.39 & 0.17 & 0.75 & 0.61 & 0.36 \\
\bottomrule
\end{tabular}
}
\end{small}
\end{table}

\textbf{Performance of Regions with Sparse POIs.}
For the NearbyPlaces in the prompt, we use the node field in the OSM data, and when there are POIs (e.g., bar, restaurant) near the coordinates, NearbyPlaces will be a variety of such buildings or stores. If there are no POIs near the coordinates, NearbyPlaces will be the names of nearby streets.
We show that LLMGeovec performs consistently in regions with extensive POI coverage (e.g., North America) and those with little coverage (e.g., Africa). 
Specifically, we utilize the continent boundaries to extract nodes belonging to North America and Africa separately and perform the five-fold cross-validation within the continents, the results of which are shown in Tab ~\ref{Table: Performance of Regions With Sparse POIs}. where LLMGeovec's results are similar in both regions.

\begin{table}[htbp]
\caption{Performance of Regions with Sparse POIs.}
\label{Table: Performance of Regions With Sparse POIs}
\centering
\begin{small}
\renewcommand{\arraystretch}{1.2} 
\setlength{\tabcolsep}{6pt}
\resizebox{0.49\textwidth}{!}{
\begin{tabular}{@{}lccc@{}}
\toprule
\textbf{Region} & \textbf{Population Density} & \textbf{Night Light Density} & \textbf{Annual Air Temperature} \\
\midrule
Africa         & 0.90                      & 0.94                         & 0.88                             \\
North America  & 0.82                       & 0.97                         & 0.92                             \\
\bottomrule
\end{tabular}
}
\end{small}
\end{table}


\subsection{LLMGeovec for LTSF}
In the previous section, we discussed how to seamlessly embed LLMGeovec into existing LTSF models. Next, we will conduct detailed experiments to verify the effectiveness of LLMGeovec using popular LTSF benchmarks and various models. Due to limited computational resources, we choose LLMGeovec (LLaMa3 8B) with the best performance in the GP to be added to the various models.

\textbf{Datasets and models.} 
Following the settings of ~\citet{zhang2024corrformer,wu2022timesnet,shao2022spatial}, we select five LTSF datasets from a wide range of domains, including Solar Energy, Global Wind, Global Temperature, Traffic flow, Delivery demand, and air quality. Several representative LTSF models are selected, including both Transformer-based and MLP-based methods. They are iTransformer \citep{liu2023itransformer}, TSMixer \citep{chen2023tsmixer}, RMLP \citep{li2023revisiting}, and Informer \citep{zhou2021informer}.

\textbf{Hyperparameters Settings.} We adapt the suggested hyperparameters in Time-Series-Library benchmark \citep{wang2024deep} for all model.

\textbf{Performances of LTSF.}
As shown in Table \ref{tab:main}, LLMGeovec can consistently improve the original performances of different models in almost all scenarios. This effect is noticeable in datasets related to both natural processes and human activities, which demonstrates the generality of LLM-based geolocation representation.

\textbf{Performances Comparisions using different geolocation embeddings.}
In this section, we compare the effects of two geolocation representations on LTSF models, and for reference, we also add learnable embeddings ~\citep{STID} (e.g., STID) with the same feature dimensions as LLMGeovec. For a fair comparison, all three methods use the same Adapter and model parameters. As shown in Tab. ~\ref{tab:Comparison of different geolocation embeddings on LTSF}, LLMGeovec, which contains richer spatial semantics, achieves the greatest improvement.

\begin{table}[htbp]
\centering
\caption{Different geolocation embeddings on LTSF.}
\label{tab:Comparison of different geolocation embeddings on LTSF}
\begin{small}
    \renewcommand{\arraystretch}{1.2} 
    \setlength{\tabcolsep}{8pt}
    \resizebox{0.9\columnwidth}{!}{
    \begin{tabular}{l|cc|cc|cc}
    \toprule
    \multirow{2}{*}{\textbf{Global Wind}} & \multicolumn{2}{c|}{\textbf{LLMGeovec (LLaMa 3 8B)}} & \multicolumn{2}{c|}{\textbf{Bert-whitening (Bert Base)}} & \multicolumn{2}{c}{\textbf{STID}} \\
    \cmidrule(lr){2-7}
    & \textbf{MSE} & \textbf{MAE} & \textbf{MSE} & \textbf{MAE} & \textbf{MSE} & \textbf{MAE} \\
    \midrule
    iTransformer & \textbf{3.560} & 1.300 & 3.563 & 1.301 & 3.575 & \textbf{1.299} \\
    \midrule
    TSMixer & \textbf{3.524} & \textbf{1.292} & 3.758 & 1.316 & 3.68 & 1.303 \\
    \midrule
    RMLP & \textbf{3.221} & \textbf{1.237} & 3.563 & 1.309 & 3.591 & 1.310 \\
    \bottomrule
    \end{tabular}}
\end{small}
\end{table}

\subsection{LLMGeovec for GSTF}
Finally, we evaluate the effectiveness of LLMGeovec in GSTF tasks and models. As with LTSF, we choose LLaMa3 8B to generate LLMGeovec.


\textbf{Datasets and models.} We select the large-scale LargeST traffic flow benchmark \citep{liu2023largest} and the LaDe demand dataset \citep{wu2023lade} for evaluations. Several competitive baselines that are widely adopted in related work are considered, including DCRNN \citep{DCRNN}, STGCN \citep{STGCN}, ASTGCN \citep{guo2019attention}, AGCRN \citep{bai2020adaptive}, GWNET \citep{GWNET}, MTGNN \citep{MTGNN}, and STID \citep{STID}. 

\textbf{Hyperparameters Settings.} We adopt the suggested hyperparameters in LargeST \citep{liu2023largest} for all models. 

\begin{table}[htbp]
\caption{Experimental results of GSTF in LargeST dataset.}
\label{tab:result_st}
\centering
\begin{small}
    \renewcommand{\multirowsetup}{\centering}
    \renewcommand{\arraystretch}{1} 
    \setlength{\tabcolsep}{5pt}
    \resizebox{1\columnwidth}{!}{
    \begin{tabular}{l|cc|cc|cc|c}
    \toprule
     \multirow{2}{*}{Models} & \multicolumn{2}{c|}{SD} & \multicolumn{2}{c|}{GLA} & \multicolumn{2}{c|}{GBA} & \multicolumn{1}{c}{IMP}\\
    \cmidrule(lr){2-8} 
    & MAE & RMSE  & MAE & RMSE & MAE & RMSE & (\%)\\
    \midrule
    HA  & 60.78 & 87.39  & 59.58 & 86.19 &  56.43 & 79.81 &--\\
    \midrule
    DCRNN  & 25.23 & 39.17 & 22.73 & 35.65 &  22.35 & 35.26 & \multirow{2}{*}{8.32\%}\\
    \cellcolor{gray!20}{+LLMGeovec}  & \boldres{18.70} & \boldres{31.36} & \boldres{21.43}  & \boldres{34.76}  & \boldres{21.69} & \boldres{34.37} & \\
    \midrule
    STGCN & 20.10 & 34.60  &  22.48 & 38.55 & 23.14 &  37.90 & \multirow{2}{*}{3.51\%} \\
    \cellcolor{gray!20}{+LLMGeovec}  & \boldres{19.83} & \boldres{33.21} & \boldres{22.03}  & \boldres{37.45} & \boldres{22.43} & \boldres{36.51} &  \\
    \midrule
    ASTGCN  & 25.13 & 39.88 & 28.44 & 44.13 & 26.15 & 40.25 & \multirow{2}{*}{7.98\%} \\
    \cellcolor{gray!20}{+LLMGeovec}  & \boldres{23.89} & \boldres{38.08} & \boldres{23.74}  & \boldres{38.27}  & \boldres{23.24} & \boldres{37.78} &\\
    \midrule
    AGCRN  & 18.45 & 34.40 & 20.61 & 36.23 & 20.55 & \underline{\boldres{33.91}} & \multirow{2}{*}{0.60\%} \\
    \cellcolor{gray!20}{+LLMGeovec}  & \boldres{18.21} & \boldres{33.82} & \underline{\boldres{19.88}}  & \boldres{35.96} & \underline{\boldres{19.77}} & 34.12 &\\
    \midrule
    GWNET  & {19.38} & {31.88}  & 21.23 & 33.68 & 20.84 & 34.58 & \multirow{2}{*}{3.92\%}\\
    \cellcolor{gray!20}{+LLMGeovec}  & \underline{\boldres{18.03}} & \boldres{30.06} & \boldres{20.29}  &  \underline{\boldres{32.62}} & \boldres{20.66} & \boldres{33.58} & \\
    \midrule    MTGNN  & 23.69 & 36.83& 23.47 & 37.68 & 23.73 & 36.01  & \multirow{2}{*}{8.09\%} \\
    \cellcolor{gray!20}{+LLMGeovec}  & \boldres{19.03} & \boldres{31.17} & \boldres{21.76}  & ~\boldres{34.58} & \boldres{22.55} &  \boldres{35.77} & \\
    \midrule
    MLP & 27.84 &  43.92 & 29.12 & 45.76 & 29.15 &45.64 & \multirow{2}{*}{26.53\%} \\
    \cellcolor{gray!20}{+LLMGeovec} & \textbf{19.00} & \underline{\textbf{30.03}}  & \boldres{21.07} & \boldres{34.56} & \boldres{21.42} & \boldres{34.92} &\\
    \bottomrule
    \end{tabular}}
\end{small}
\end{table}

\textbf{Performances in GSTF.} Tab. \ref{tab:result_st} reports the evaluation results on the LargeST benchmark. Mainstream GNN-based models can benefit from the incorporation of LLMGeovec. This clearly shows that LLMGeovec is able to complement the spatial relationships captured by GNNs with the rich geographic knowledge of LLMs.
Surprisingly, the vanilla MLP model equipped with LLMGeovec can achieve comparable performance to the GNN counterparts. This suggests that LLMGeovec can even be used as an alternative to GNNs to provide geographic correlation for temporal models.

\textbf{Overhead of LLMGeovec.}
For the GTSF and LTSF tasks, we concatenate the d-dimensional LLMGeovec to the original feature vector while reducing the dimensionality of the original feature vector by d. This way, the overall dimensionality remains unchanged. The only additional parameters introduced are the LLM Adapter (A two-layer MLP that uses Leaky\_Relu in the middle), which is used to reduce the input dimensionality of the LLMGeovec. As shown in ~\ref{tab:Comparison of STGCN with and without LLMGeovec}, our model introduces only an acceptable number of additional parameters and maintains computational efficiency comparable to the original model.

\begin{table}[htbp]
\centering
\caption{Ovearhead of LLMGeovec.}
\label{tab:Comparison of STGCN with and without LLMGeovec}
\begin{small}
    \renewcommand{\arraystretch}{1.2} 
    \setlength{\tabcolsep}{8pt}
    \resizebox{1\columnwidth}{!}{
    \begin{tabular}{l|c|c|c}
    \toprule
    \textbf{Model} & \textbf{Parameters} & \textbf{Running Speed} & \textbf{GPU Memory} \\
    \midrule
    STGCN w/o LLMGeovec & 3482K & 3.62 s/it & 3.5 GB \\
    \midrule
    STGCN w/ LLMGeovec & 3351K & 3.58 s/it & 3.5 GB \\
    \bottomrule
    \end{tabular}
    }
\end{small}
\end{table}

\begin{figure}[!htbp]
  \centering
  \captionsetup{skip=1pt}
  \includegraphics[width=1\columnwidth]{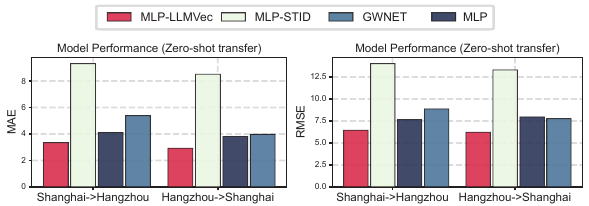}
  \caption{Comparsion in zero-shot transfer scenarios.}
  \label{fig:zero_shot_compare}
\end{figure}

\textbf{Performances in zero-shot scenarios.}
In addition to enhancing various models in full training scenarios for GSTF, LLMGeovec also has the potential for enhancing zero-shot transfer. We compare the performance of the learnable node embedding (STID) introduced by \citet{shao2022spatial} and LLMGeovec in zero-shot scenarios. As a reference, we also test the transferability of the baseline GWNET and MLP. The LaDe data is adopted for this experiment.
It is evident from Fig. \ref{fig:zero_shot_compare} that when models are transferred to a new region in a zero-shot scenario, the learnable embedding harms the performance of MLP significantly because the embedding has adapted to the source data with specific patterns. In contrast, universal LLMGeovec can be generalized to other regions without any adjustment. This indicates that LLMGeovec features intrinsic geolocation knowledge that is generalizable for different regions. In addition, more advanced techniques such as test-time adaptation in graphs can be applied to achieve better few-shot performance \cite{sun2024program}.

\section{Conclusion and Future Work}
The acquisition of universal geolocation representations to improve downstream tasks has been a long-standing pursuit. This paper presents our first attempt to utilize recent advanced LLMs to extract such representations. By the merit of the geospatial knowledge within LLMs, the extracted embedding from the pre-activated layer achieves global coverage and serves as a generic enhancer for spatio-temporal learning. We demonstrate the effectiveness of embedding in various tasks, including GP, LTSF, and GSTF. Empirical results indicate that LLMGeovec can improve the performances of various models simply by incorporating it into the input (i.e., feature concatenation).
In future work, we are interested to see if larger LLMs (e.g., LLaMa3 70B) can further improve the quality of LLMGeovec. We can adopt LLMGeovec in more challenging spatio-temporal learning tasks, such as spatio-temporal imputation \citep{nie2024imputeformer,yuan2022stgan} and traffic flow generation ~\citep{wu2020spatiotemporal}.  It is also interesting to explore the possibility of integrating LLMGeovec into pre-trained foundational models for unified spatio-temporal learning \citep{jin2023time,yuan2024unist,zhang2024large}.



\section*{Acknowledgments}
The work described in this paper was supported by grants from the Research Grants Council of the Hong Kong Special Administrative Region, China (Project No. PolyU/15206322 and PolyU/15227424),

\bibliography{aaai25}

\clearpage

\clearpage

\appendix


\section{Appendices}


\subsection{Descriptions of datasets and models in GP}

\subsubsection{Datasets}
We utilized a variety of global-scale datasets to inform our models. The datasets are listed below with their key characteristics:
\begin{itemize}
    \item \textbf{Annual Air Temperature:} Mean annual daily mean air temperature data. \\ \path{CHELSA_bio1_1981-2010_V.2.1.tif}
    
    \item \textbf{Annual Precipitation:} Mean annual accumulated precipitation amount. \\ \path{CHELSA_bio12_1981-2010_V.2.1.tif}
    
    \item \textbf{Monthly Climate Moisture:} Average monthly climate moisture index. \\ \path{CHELSA_cmi_mean_1981-2010_V.2.1.tif}
    
    \item \textbf{Population Density:} GeoLLM aggregates WorldPop population data at 1km resolution, employing importance sampling by population size. \\ \path{ppp_2020_1km_Aggregated.tif}
    
    \item \textbf{Nighttime Light Intensity:} Satellite images capturing nighttime luminosity from VIIRS with a 500-meter resolution. \\ \path{VNL_npp_2023_global_vcmslcfg_v2_c202402081600.cvg.dat.tif}

    \item \textbf{Human Modification Terrestrial:} A cumulative metric of human modification of terrestrial lands at a 1-km resolution, modeled from 13 anthropogenic stressors. \\ \path{lulc-human-modification-terrestrial-systems_geographic.tif}
    
    \item \textbf{Global Gridded Relative Deprivation:} Index of multidimensional deprivation and poverty, ranging from 0 (lowest) to 100 (highest), at 30 arc-second (~1 km) resolution. \\ \path{povmap-grdi-v1.tif}
    
    \item \textbf{Other Indicators:} Additional indicators include the Ratio of Built-up Area to Non-built Up Area, Child Dependency Ratio, and Subnational Human Development Index. \\ \path{povmap-grdi-v1_BUILT.tif}, \path{povmap-grdi-v1_CDR_CopyRaster.tif}, \path{povmap-grdi-v1_SHDI.tif}
    
    \item \textbf{Infant Mortality Rates:} Subnational Infant Mortality Rate estimates for 234 countries and territories. \\ \path{povmap_global_subnational_infant_mortality_rates_v2_01.tif}

    \item \textbf{SustainBench Indicators:} Asset index, sanitation index, and women's BMI collected from Demographic and Health Surveys across 48 countries. \\ \path{dhs_asset_index.tif}, \path{dhs_sanitation_index.tif}, \path{dhs_women_bmi.tif}
\end{itemize}

For country-scale and city-scale datasets, we used the datasets provided by MapillaryGCN and HKGL, respectively. The MapillaryGCN dataset records poverty rate, population density, and women's BMI across 6,000 communities in India, as detailed in the original article. The HKGL dataset captures population density, education level, income level, and crime rate across 1,500 census tracts in NYC.

\subsubsection{Models}
We evaluate several baseline models, as detailed below:

\begin{itemize}
    \item \textbf{Image Supervised Learning}: Trains a ResNet34 pre-trained on ImageNet1k to predict cluster-specific indicators from street view images.
    \item \textbf{Object Counts}: Utilizes object detection outputs from street view images, followed by MLPs for indicator prediction.
    \item \textbf{MapillaryGCN}: Combines street view images and object detection results with a GCN for feature aggregation and prediction.
    \item \textbf{Node2Vec}: Employs random walks to learn node embeddings using skip-gram models.
    \item \textbf{GCN}: Aggregates information from neighboring nodes for embedding learning.
    \item \textbf{GAT}: Utilizes attention mechanisms to differentially weight information from neighboring nodes during aggregation.
    \item \textbf{ZE-Mob}: Leverages the co-occurrence of origin-destination locations to learn location embeddings from mobility flow data.
    \item \textbf{MGFN}: Fuses mobility graphs with similar patterns, then learns location embeddings using a multi-level attention mechanism.
    \item \textbf{MV-PN}: Constructs multi-view POI-POI networks per location and learns embeddings through an encoder-decoder framework.
    \item \textbf{HDGE}: Jointly learns location embeddings from both spatial and flow graphs.
    \item \textbf{HUGAT}: Defines meta-paths to capture semantics in LBSN, applying a heterogeneous graph attention network for embedding learning.
    \item \textbf{MVURE}: Models various location correlations using different graphs, with a joint learning module for location embeddings.
    \item \textbf{HKGL}: Implements a hierarchical KG learning model, using LBKG for global knowledge distillation and sub-KGs for domain-specific knowledge capture.
    \item \textbf{Bert-whitening}: In Bert, sentence representations are obtained by averaging the vectors of individual words, on the basis of which the isotropy of sentence representations is enhanced by whitening operations.
\end{itemize}

\subsection{Datasets and Models in LTSF}
\subsubsection{Datasets}
We evaluate the proposed models on a diverse set of datasets, each characterized by distinct temporal patterns:

\begin{itemize}
    \item \textbf{Global Wind, Global Temp:} This dataset, provided by Corrformer, originates from the National Centers for Environmental Information (NCEI). It encompasses hourly averaged wind speeds and temperatures from 3,850 global stations, spanning from January 1, 2019, to December 31, 2020.
    \item \textbf{Solar Energy:} Consists of electricity generation data from 137 solar stations in Alabama, recorded at 15-minute intervals.
    \item \textbf{Demand-SH:} A delivery demand dataset from Shanghai, provided by LaDe, comprising 96,000 trajectories over a 6-month period.
    \item \textbf{Air Quality:} Includes air quality measurements from 437 cities across China.
    \item \textbf{Traffic-SD:} Contains traffic flow data at 716 nodes in San Diego, recorded every 5 minutes from January 1, 2017, to December 31, 2021.
\end{itemize}

\subsubsection{Models}
We benchmark the performance of the following models:

\begin{itemize}
    \item \textbf{iTransformer:} Applies the attention mechanism and a feed-forward network to inverted dimensions. Specifically, time points of individual series are embedded into variate tokens, which are utilized by the attention mechanism to capture multivariate correlations. Concurrently, the feed-forward network is employed on each variate token to learn non-linear representations.
    \item \textbf{TSMixer:} Based on mixing operations across both time and feature dimensions, TSMixer efficiently extracts relevant information.
    \item \textbf{RMLP:} Utilizes RevIN (reversible normalization) and Channel Independence (CI) to enhance overall forecasting performance.
    \item \textbf{Informer:} Investigates sparsity in the self-attention mechanism and proposes an efficient Transformer architecture tailored for LTSF.
    \item \textbf{STID}: Identifies nodes and time slots using learnable embedding.
\end{itemize}

\subsection{Descriptions of datasets and models in GSTF}

\subsubsection{Datasets}
We utilize the following datasets:
\begin{itemize}
    \item \textbf{LargeST-SD, GLD, GBA:} These datasets are provided by LargeST and include traffic flow data spanning five years from three cities, covering approximately 8,600 nodes.
    \item \textbf{Delivery Demand-ShangHai, Hangzhou:} Provided by LaDe, these datasets contain detailed package information, such as location and time requirements. They also include event logs documenting courier activities, such as task acceptance and completion.
\end{itemize}

\subsubsection{Models}
Our research employs several models:
\begin{itemize}
  \item \textbf{DCRNN:} This model leverages bidirectional random walks on graphs to capture spatial dependencies and uses an encoder-decoder architecture with scheduled sampling to address temporal dependencies.
  \item \textbf{STGCN:} STGCN employs a fully convolutional structure to address graph-based problems, enabling faster training and reduced model complexity.
  \item \textbf{ASTGCN:} Integrating gated recurrent units with adaptive graph convolutional networks, ASTGCN dynamically learns graph structures while capturing spatial and local temporal relationships.
  \item \textbf{AGCRN:} AGCRN introduces two adaptive modules that enhance GCN capabilities: a Node Adaptive Parameter Learning (NAPL) module for node-specific pattern learning and a Data Adaptive Graph Generation (DAGG) module for inferring automatic inter-dependencies among traffic series.
  \item \textbf{GWNET:} GWNET develops a novel adaptive dependency matrix, learning through node embedding to capture the hidden spatial dependencies in data.
  \item \textbf{MTGNN:} This model features a graph learning module for extracting uni-directed variable relations, a mix-hop propagation layer, and a dilated inception layer, enhancing both spatial and temporal dependency capture within the time series.
\end{itemize}

\subsection{Full results of LTSF}
Full results of LTSF are shown in Table \ref{tab:ltsf_appendix}.
We can see that LLMGeovec leads to direct improvement in various LSTF models and datasets with different prediction lengths.

\begin{table*}[ht]
  \caption{Full Results of the LTSF  benchmarks. We report the forecast error of different models under different prediction lengths. IMP shows the average percentage of MSE improvement of LLMGeovec.}
  \label{tab:ltsf_appendix}
  \vskip -0.0in
  \vspace{0pt}
  \renewcommand{\arraystretch}{1} 
  \centering
  \resizebox{0.8\textwidth}{!}{
  \begin{threeparttable}
  \begin{small}
  \renewcommand{\multirowsetup}{\centering}
  \setlength{\tabcolsep}{2pt}
  \begin{tabular}{c|c|cc>{\columncolor{gray!20}}c>{\columncolor{gray!20}}c|cc>{\columncolor{gray!20}}c>{\columncolor{gray!20}}c|cc>{\columncolor{gray!20}}c>{\columncolor{gray!20}}c|cc>{\columncolor{gray!20}}c>{\columncolor{gray!20}}c|c}
    \toprule
    \multicolumn{2}{c}{\multirow{2}{*}{Models}}  &
    \multicolumn{2}{|c}{\rotatebox{0}{{iTransformer}}} &
    \multicolumn{2}{c|}{\rotatebox{0}{{w/ LLMGeovec}}} &
    \multicolumn{2}{c}{\rotatebox{0}{{TSMixer}}} &
    \multicolumn{2}{c|}{\rotatebox{0}{{w/ LLMGeovec}}} &
    \multicolumn{2}{c}{\rotatebox{0}{{RMLP}}}&
    \multicolumn{2}{c}{\rotatebox{0}{{w/ LLMGeovec}}} &
    \multicolumn{2}{|c}{\rotatebox{0}{{Informer}}} &
    \multicolumn{2}{c|}{\rotatebox{0}{{w/ LLMGeovec}}} &
    \multicolumn{1}{c}{\rotatebox{0}{IMP}}\\
    \cmidrule(lr){3-19} 
    \multicolumn{2}{c}{Metric}   & \multicolumn{1}{|c}{{MSE}} & \multicolumn{1}{c}{MAE}  & \multicolumn{1}{c}{MSE} & \multicolumn{1}{c|}{MAE}  & \multicolumn{1}{c}{MSE} & \multicolumn{1}{c}{MAE}  & \multicolumn{1}{c}{MSE} & \multicolumn{1}{c|}{MAE}  & \multicolumn{1}{c}{MSE} & \multicolumn{1}{c}{MAE}  & \multicolumn{1}{c}{MSE} & \multicolumn{1}{c|}{MAE} & \multicolumn{1}{c}{MSE} & \multicolumn{1}{c}{MAE} & \multicolumn{1}{c}{MSE} & \multicolumn{1}{c|}{MAE}  & $\%$\\
    \toprule
     \multirow{5}{*}{\rotatebox{90}{{Global Wind}}} 
    &  {24}  &{3.812}  &{1.440}  &{\boldres{3.222}}  &{\boldres{1.237}}  &{3.583}  &{1.313}  &{\boldres{3.524}}  &{\boldres{1.292}}  &{3.873}  &{1.356}  &{\boldres{3.562}}  &{\boldres{1.300}}  &{4.708}  &{1.532}  &{\boldres{4.683}}  &{\boldres{1.524}}  &{15.47\%}    
    \\
    & {48} &{4.441}  &{1.456}  &{\boldres{3.851}}  &{\boldres{1.354}}  &{4.191}  &{1.414}  &{\boldres{4.105}}  &{\boldres{1.391}}  &{4.521}  &{1.476}  &{\boldres{4.160}}  &{\boldres{1.411}}  &{4.842}  &{1.558}  &{\boldres{4.749}}  &{\boldres{1.537}}  &{13.29\%}    
    \\
    & {96}  &{4.904}  &{1.553}  &{\boldres{4.298}}  &{\boldres{1.435}}  &{4.670}  &{1.484}  &{\boldres{4.416}}  &{\boldres{1.479}}  &{4.997}  &{1.559}  &{\boldres{4.434}}  &{\boldres{1.457}}  &{5.081}  &{1.623}  &{\boldres{4.960}}  &{\boldres{1.603}}  &{12.36\%}    
    \\
    & {168} &{5.171}  &{1.602}  &{\boldres{4.546}}  &{\boldres{1.493}}  &{4.598}  &{1.486}  &{\boldres{4.484}}  &{\boldres{1.466}}  &{5.260}  &{1.607}  &{\boldres{4.564}}  &{\boldres{1.488}}  &{4.987}  &{\boldres{1.591}}  &{\boldres{4.983}}  &{1.600}  &{12.08\%}    
    \\
    \cmidrule(lr){2-19}
    & {Avg}  &{4.582}  &{1.513}  &{\boldres{3.979}}  &{\boldres{1.380}}  &{4.261}  &{1.424}  &{\boldres{4.132}}  &{\boldres{1.407}}  &{4.905}  &{1.498}  &{\boldres{4.180}}  &{\boldres{1.414}}  &{4.905}  &{1.576}  &{\boldres{4.844}}  &{\boldres{1.566}}  &{13.30\%}    
    \\
    \midrule
    \multirow{5}{*}{\rotatebox{90}{{Global Temp}}} 
    &  {24}  &{9.000}  &{2.048}  &{\boldres{8.808}}  &{\boldres{2.040}}  &{7.564}  &{1.945}  &{\boldres{6.927}}  &{\boldres{1.855}}  &{8.555}  &{1.997}  &{\boldres{8.000}}  &{\boldres{1.931}}  &{\boldres{15.200}}  &{\boldres{2.893}}  &{16.226}  &{3.014}  &{2.13\%}    
    \\
    & {48}  &{12.447}  &{2.481}  &{\boldres{11.346}}  &{\boldres{2.383}}  &{10.406}  &{2.324}  &{\boldres{10.290}}  &{\boldres{2.304}}  &{11.752}  &{2.392}  &{\boldres{11.086}}  &{\boldres{2.332}}  &{16.494}  &{3.038}  &{\boldres{16.181}}  &{\boldres{2.997}}  &{8.85\%}    
    \\
    & {96}  &{16.295}  &{2.916}  &{\boldres{15.623}}  &{\boldres{2.869}}  &{13.738}  &{2.696}  &{\boldres{12.790}}  &{\boldres{2.632}}  &{15.293}  &{2.782}  &{\boldres{14.263}}  &{\boldres{2.703}}  &{\boldres{19.333}}  &{3.311}  &{19.334}  &{\boldres{3.309}}  &{4.12\%}    
    \\
    & {168}  &{19.076}  &{3.169}  &{\boldres{18.003}}  &{\boldres{3.114}}  &{16.433}  &{2.955}  &{\boldres{15.768}}  &{\boldres{2.900}}  &{18.187}  &{3.061}  &{\boldres{16.752}}  &{\boldres{2.954}}  &{\boldres{22.453}}  &{\boldres{3.596}}  &{22.814}  &{3.625}  &{5.63\%}    
    \\
    \cmidrule(lr){2-19}
    & {Avg}  &{13.079}  &{2.653}  &{\boldres{11.945}}  &{\boldres{2.601}}  &{12.035}  &{2.480}  &{\boldres{11.441}}  &{\boldres{2.398}}  &{13.447}  &{2.558}  &{\boldres{12.525}}  &{\boldres{2.480}}  &{18.370}  &{3.209}  &{\boldres{18.639}}  &{\boldres{3.234}}  &{5.19\%}    
   
    \\
    \midrule
    \multirow{5}{*}{\rotatebox{90}{{Solar Energy}}} 
    &  {96}  &{0.203} &{0.237} & {\boldres{0.180}} & {\boldres{0.213}} &{0.222} &{0.281} & {\boldres{0.199}} & {\boldres{0.281}}  &{0.233} &{0.296} & {\boldres{0.213}} & {\boldres{0.271}} &{0.236} &{ \boldres{0.279}} & {\boldres{0.227}} & {0.289} &{11.33\%} 
    \\
    & {192}  &{0.233} &{\boldres{0.261}} & {\boldres{0.217}} & {0.289} &{0.261} &{\boldres{0.301}} & {\boldres{0.229}} & {0.308}  &{0.260} &{0.316} & {\boldres{0.239}} & {\boldres{0.291}} &{0.227} &{ 0.287} & {\boldres{0.303}} & {\boldres{0.332}} & {6.87\%} 
    \\
    & {336}  &{0.248} &{\boldres{0.273}} & {\boldres{0.216}} & {0.291} &{0.271} &{0.299} & {\boldres{0.206}} & {\boldres{0.270}}  &{0.276} &{0.323} & {\boldres{0.244}} & {\boldres{0.292}} &{\boldres{0.262}} &{ 0.310} & {0.254} & {\boldres{0.309}} & {12.90\%} 
    \\
    & {720}  &{0.249} &{\boldres{0.275}} & {\boldres{0.211}} & {0.277} &{0.267} &{\boldres{0.293}} & {\boldres{0.243}} & {0.306}  &{0.273} &{0.316} & {\boldres{0.244}} & {\boldres{0.288}} &{0.329} &{ 0.355} & {\boldres{0.259}} & {\boldres{0.323}} & {15.26\%} 
    \\
    \cmidrule(lr){2-19}
    & {Avg}  &{0.233}  &{0.262}  &{\boldres{0.206}}  &{\boldres{0.265}}  &{0.255}  &{0.294}  &{\boldres{0.219}}  &{\boldres{0.289}}  &{0.261}  &{0.313}  &{\boldres{0.235}}  &{\boldres{0.286}}  &{0.264}  &{0.308}  &{\boldres{0.263}}  &{0.313}  &{11.59\%}  
  
    \\
    \midrule
    \multirow{5}{*}{\rotatebox{90}{{Demand-SH}}} 
    &  {48}  &{0.238}  &{0.256}  &{\boldres{0.233}}  &{\boldres{0.253}}  &{0.259}  &{0.282}  &{\boldres{0.246}}  &{\boldres{0.259}}  &{0.244}  &{0.271}  &{\boldres{0.227}}  &{\boldres{0.242}}  &{0.474}  &{0.401}  &{\boldres{0.385}}  &{\boldres{0.359}}  &{2.10\%}    
    \\
    & {96}  &{0.291}  &{0.282}  &{\boldres{0.285}}  &{\boldres{0.281}}  &{0.315}  &{0.311}  &{\boldres{0.301}}  &{\boldres{0.290}}  &{0.302}  &{0.306}  &{\boldres{0.279}}  &{\boldres{0.268}}  &{0.588}  &{0.479}  &{\boldres{0.489}}  &{\boldres{0.418}}  &{2.06\%}    
    \\
    & {168}  &{0.360}  &{0.314}  &{\boldres{0.351}}  &{\boldres{0.313}}  &{0.383}  &{0.348}  &{\boldres{0.362}}  &{\boldres{0.319}}  &{0.373}  &{0.343}  &{\boldres{0.345}}  &{\boldres{0.299}}  &{1.028}  &{0.686}  &{\boldres{0.537}}  &{\boldres{0.461}}  &{2.50\%}   
    \\
    & {360}  &{0.434}  &{0.340}  &{\boldres{0.420}}  &{\boldres{0.339}}  &{0.461}  &{0.386}  &{\boldres{0.433}}  &{\boldres{0.350}}  &{0.459}  &{0.384}  &{\boldres{0.420}}  &{\boldres{0.332}}  &{1.475}  &{0.904}  &{\boldres{0.704}}  &{\boldres{0.525}}  &{3.23\%}   
    \\
    \cmidrule(lr){2-19}
    & {Avg} & {0.331}  &{0.298}  &{\boldres{0.322}}  &{\boldres{0.297}}  &{0.355}  &{0.332}  &{\boldres{0.336}}  &{\boldres{0.305}}  &{0.345}  &{0.326}  &{\boldres{0.318}}  &{\boldres{0.286}}  &{0.896}  &{\boldres{0.618}}  &{\boldres{0.779}}  &{0.666}  &{2.47\%} 
   
    \\
    \midrule
    \multirow{5}{*}{\rotatebox{90}{{Air Quality}}} 
    & {6}  &{1.155}  &{0.467}  &{\boldres{1.126}}  &{\boldres{0.465}}  &{1.289}  &{0.507}  &{\boldres{1.259}}  &{\boldres{0.505}}  &{1.235}  &{0.495}  &{\boldres{1.158}}  &{\boldres{0.468}}  &{3.542}  &{0.817}  &{\boldres{2.880}}  &{\boldres{0.713}}  &{2.51\%}   
    \\
    & {12}  &{1.672}  &{0.593}  &{\boldres{1.589}}  &{\boldres{0.567}}  &\boldres{1.775}  &{0.606}  &{{1.787}}  &{\boldres{0.602}}  &{1.629}  &{0.583}  &{\boldres{1.610}}  &{\boldres{0.576}}  &{3.409}  &{0.807}  &{\boldres{3.087}}  &{\boldres{0.756}}  &{4.97\%}   
    \\
    & {24}  &{2.155}  &{0.683}  &{\boldres{2.081}}  &{\boldres{0.673}}  &{2.333}  &{0.727}  &{\boldres{2.196}}  &{\boldres{0.701}}  &{2.048}  &{0.671}  &{\boldres{2.028}}  &{\boldres{0.664}}  &{4.859}  &{0.955}  &{\boldres{3.236}}  &{\boldres{0.792}}  &{3.44\%}   
    \\
    & {48} &{2.707}  &{0.781}  &{\boldres{2.628}}  &{\boldres{0.770}}  &{2.875}  &{0.819}  &{\boldres{2.713}}  &{\boldres{0.791}}  &{2.517}  &{0.757}  &{\boldres{2.483}}  &{\boldres{0.744}}  &{3.524}  &{0.878}  &{\boldres{3.228}}  &{\boldres{0.824}}  &{2.92\%}   
    \\
    \cmidrule(lr){2-19}
    & {Avg}  &{1.922}  &{0.631}  &{\boldres{1.856}}  &{\boldres{0.619}}  &{2.068}  &{0.665}  &{\boldres{1.989}}  &{\boldres{0.650}}  &{1.857}  &{0.627}  &{\boldres{1.820}}  &{\boldres{0.613}}  &{3.584}  &{0.864}  &{\boldres{2.858}}  &{\boldres{0.771}}  &{3.46\%}  
  
    \\
    \midrule
    \multirow{5}{*}{\rotatebox{90}{{Traffic-SD}}} 
    & {96}  &{0.104}  &{0.195}  &{\boldres{0.080}}  &{\boldres{0.176}}  &{0.090}  &{0.183}  &{\boldres{0.086}}  &{\boldres{0.173}}  &{0.175}  &{0.264}  &{\boldres{0.140}}  &{\boldres{0.237}}  &{0.167}  &{0.267}  &{\boldres{0.124}}  &{\boldres{0.225}}  &{23.08\%}   
    \\
    & {192}  &{0.139}  &{0.229}  &{\boldres{0.104}}  &{\boldres{0.202}}  &{0.111}  &{0.207}  &{\boldres{0.098}}  &{\boldres{0.191}}  &{0.217}  &{0.305}  &{\boldres{0.174}}  &{\boldres{0.270}}  &{0.207}  &{0.304}  &{\boldres{0.140}}  &{\boldres{0.239}}  &{ 25.18\%}   
    \\
    & {336}  &{0.167}  &{0.252}  &{\boldres{0.128}}  &{\boldres{0.221}}  &{0.138}  &{0.234}  &{\boldres{0.120}}  &{\boldres{0.215}}  &{0.229}  &{0.320}  &{\boldres{0.189}}  &{\boldres{0.286}}  &{0.224}  &{0.321}  &{\boldres{0.167}}  &{\boldres{0.266}}  &{23.35\%}   
    \\
    & {720}  &{0.134}  &{0.223}  &{\boldres{0.112}}  &{\boldres{0.205}}  &{0.124}  &{0.225}  &{\boldres{0.115}}  &{\boldres{0.209}}  &{0.201}  &{0.293}  &{\boldres{0.168}}  &{\boldres{0.264}}  &{0.199}  &{0.298}  &{\boldres{0.176}}  &{\boldres{0.285}}  &{16.42\%}   
    \\
    \cmidrule(lr){2-19}
    & {Avg}  &{0.136}  &{0.225}  &{\boldres{0.106}}  &{\boldres{0.201}}  &{0.116}  &{0.212}  &{\boldres{0.105}}  &{\boldres{0.197}}  &{0.205}  &{0.296}  &{\boldres{0.168}}  &{\boldres{0.264}}  &{0.199}  &{0.298}  &{\boldres{0.152}}  &{\boldres{0.254}}  &{22.01\%}  
  
    \\
    \bottomrule
  \end{tabular}
    \end{small}
  \end{threeparttable}
}
\end{table*}

\end{document}